\documentclass{article}

\usepackage{arxiv}

\usepackage[utf8]{inputenc}
\usepackage[T1]{fontenc}
\usepackage{hyperref}
\usepackage{url}
\usepackage{booktabs}
\usepackage{amsfonts}
\usepackage{amsmath}
\usepackage{amssymb}
\usepackage{microtype}
\usepackage{graphicx}
\usepackage{multirow}
\usepackage{array}
\usepackage{xcolor}
\usepackage{natbib}
\usepackage{doi}
\usepackage{float}
\usepackage{caption}
\usepackage{subcaption}
\usepackage{enumitem}
\usepackage{bm}

\title{Task-Conditioned Routing Signatures in Sparse Mixture-of-Experts Transformers}

\author{
MSR Avinash \\
Independent Researcher | Asthra Labs \\
\texttt{aviinashh.mynampati@gmail.com}
}

\renewcommand{\shorttitle}{Task-Conditioned Routing Signatures}

\hypersetup{
pdftitle={Task-Conditioned Routing Signatures in Sparse Mixture-of-Experts Transformers},
pdfsubject={Machine Learning, Interpretability, Sparse Transformers},
pdfauthor={MSR Avinash},
pdfkeywords={Mixture of Experts, MoE, Routing, Interpretability, Sparse Transformers, OLMoE},
}

\begin{document}

\maketitle

\begin{abstract}
Sparse Mixture-of-Experts (MoE) architectures enable efficient scaling of large language models through conditional computation, yet the routing mechanisms responsible for expert selection remain poorly understood. In this work, we introduce \emph{routing signatures}, a vector representation summarizing expert activation patterns across layers for a given prompt, and use them to study whether MoE routing exhibits task-conditioned structure. Using OLMoE-1B-7B-0125-Instruct as an empirical testbed, we show that prompts from the same task category induce highly similar routing signatures, while prompts from different categories exhibit substantially lower similarity. Within-category routing similarity ($0.8435 \pm 0.0879$) significantly exceeds across-category similarity ($0.6225 \pm 0.1687$), corresponding to Cohen's $d = 1.44$. A logistic regression classifier trained solely on routing signatures achieves $92.5\% \pm 6.1\%$ cross-validated accuracy on four-way task classification. To ensure statistical validity, we introduce permutation and load-balancing baselines and show that the observed separation is not explained by sparsity or balancing constraints alone. We further analyze layer-wise signal strength and low-dimensional projections of routing signatures, finding that task structure becomes increasingly apparent in deeper layers. These results suggest that routing in sparse transformers is not merely a balancing mechanism, but a measurable task-sensitive component of conditional computation. We release \textsc{MoE-Xray}, a lightweight toolkit for routing telemetry and analysis.
\end{abstract}


\section{Introduction}

Sparse Mixture-of-Experts (MoE) transformers have emerged as a compelling architecture for scaling language models while controlling inference cost. Instead of activating the full parameter set for each token, MoE models use a learned router to select a small subset of experts, allowing them to increase total model capacity without proportional increases in active computation. This conditional computation paradigm has been central to the success of large sparse architectures such as GShard, Switch Transformers, and more recent open MoE systems \citep{lepikhin2020gshard,fedus2022switch,muennighoff2024olmoe}.

Despite the architectural importance of routing, its internal behavior remains poorly characterized. Existing work has largely focused on training stability, scaling behavior, and balancing losses \citep{lepikhin2020gshard,fedus2022switch,zhou2022expertchoice}. In contrast, relatively little work studies routing \emph{as an object of analysis}: a structured signal that may reveal how sparse models allocate computation across tasks. This gap matters for three reasons.

First, routing is central to \textbf{interpretability}. If different tasks systematically activate different experts, then routing patterns may provide a tractable view into how sparse transformers organize computation. Second, routing matters for \textbf{debugging and monitoring}. Abnormal routing patterns may signal expert collapse, drift, or degradation in deployed MoE systems. Third, routing may reveal whether sparse models implement different \textbf{computation pathways} for different tasks, a question closely tied to the broader study of modularity in neural systems.

In this paper we investigate whether routing behavior contains task-conditioned structure. Specifically, we ask whether prompts from different categories induce statistically distinguishable expert activation patterns. To do so, we introduce the concept of a \emph{routing signature}: a compact representation of expert usage frequencies across layers for a given prompt. By comparing routing signatures across prompts, we can measure whether routing patterns cluster by task.

Our empirical setting is intentionally simple and controlled. We evaluate OLMoE-1B-7B-0125-Instruct across 80 prompts spanning four categories: code, math, story, and factual question-answering. We then ask three core questions:

\begin{enumerate}[leftmargin=*]
    \item Do routing signatures cluster by task category?
    \item Does this clustering exceed what would be expected under random or purely load-balanced routing?
    \item Do routing signatures contain enough information to predict task identity?
\end{enumerate}

Our results answer all three questions positively. Prompts from the same category exhibit much higher routing similarity than prompts from different categories. This separation exceeds both permutation and load-balancing baselines. Furthermore, a simple linear classifier trained only on routing signatures achieves over 92\% accuracy on task classification. Together, these findings suggest that routing is not merely a balancing mechanism, but a measurable task-sensitive component of sparse transformer computation.

\subsection{Contributions}

This paper makes the following contributions:

\begin{enumerate}[leftmargin=*]
    \item We introduce \textbf{routing signatures}, a compact representation of expert activation patterns across layers.
    \item We provide a \textbf{statistical framework} for comparing routing patterns across prompts and tasks.
    \item We demonstrate \textbf{strong task-conditioned clustering} of routing signatures in OLMoE.
    \item We validate this effect against \textbf{permutation and load-balancing baselines}.
    \item We show that routing signatures support \textbf{high-accuracy linear classification} of task category.
    \item We release \textbf{\textsc{MoE-Xray}}, a lightweight toolkit for routing telemetry and analysis.
\end{enumerate}

\section{Background}

\subsection{Mixture-of-Experts Routing}

In a standard dense transformer, each token at layer $\ell$ is processed by a single shared feedforward block. In an MoE transformer, this block is replaced by a set of experts $\{f_1, \dots, f_E\}$ together with a learned router. The router computes a distribution over experts for each token representation and selects a subset for execution.

Let $h_{\ell,t} \in \mathbb{R}^d$ denote the hidden state of token $t$ at layer $\ell$. The router computes expert selection probabilities as

\begin{equation}
p(e \mid h_{\ell,t}) = \text{softmax}(W_r h_{\ell,t}),
\end{equation}

where $W_r \in \mathbb{R}^{E \times d}$ is the router projection matrix and $e \in \{1,\dots,E\}$ indexes experts.

In top-$k$ routing, the $k$ experts with highest probability are selected. Only these experts are active for token $t$ at layer $\ell$. The outputs of the selected experts are then combined, typically using router-derived mixture weights.

This mechanism enables \emph{conditional computation}: different tokens can be processed by different experts, allowing the model to allocate specialized capacity dynamically.

\subsection{Conditional Computation and Task Structure}

Because routing decisions depend on hidden representations, and hidden representations depend on both token identity and context, routing behavior is potentially sensitive to the structure of the input task. A code-generation prompt, for example, may produce systematically different hidden states from a creative writing prompt, leading the router to select different experts.

If this is true, then routing patterns should not be random. Instead, they should contain structure reflecting the task distribution over hidden states. This motivates analyzing routing not merely as a training mechanism, but as a statistical signature of internal computation.

\subsection{Prior Work}

Previous MoE research has primarily emphasized scaling efficiency, balancing objectives, and training dynamics. GShard introduced large-scale sharded MoE training \citep{lepikhin2020gshard}. Switch Transformers simplified routing to single-expert dispatch while emphasizing scalability and load balancing \citep{fedus2022switch}. Expert Choice Routing explored alternative dispatch strategies with stronger expert utilization guarantees \citep{zhou2022expertchoice}. More recent open models such as OLMoE have made sparse architectures more accessible for empirical analysis \citep{muennighoff2024olmoe}.

What has been less explored is whether routing itself forms a structured signal that can be compared across prompts, tasks, and models. Our work is aimed at this gap.

\section{Routing Signature Framework}

\subsection{Routing Events}

We define a \emph{routing event} as the activation of expert $e$ for token $t$ at layer $\ell$:

\begin{equation}
(\ell, t, e).
\end{equation}

A single prompt produces many routing events across tokens and layers. Together, these events form the raw routing trace of the prompt.

\subsection{Routing Signatures}

To compare prompts, we summarize routing traces using \emph{routing signatures}. Let $A_{\ell,e}(x)$ denote the number of times expert $e$ is activated at layer $\ell$ for prompt $x$. We define the layer-wise routing signature as

\begin{equation}
s_{\ell,e}(x)=
\frac{A_{\ell,e}(x)}
{\sum_{e'} A_{\ell,e'}(x)}.
\end{equation}

This normalizes expert activation counts within each layer, producing a distribution over experts for every layer. Concatenating across all layers yields a full routing signature:

\begin{equation}
s(x) \in \mathbb{R}^{L \cdot E}.
\end{equation}

For the model used in this work, $L=16$ and $E=64$, so each routing signature has dimensionality $1024$.

Intuitively, a routing signature acts as a fingerprint of how a prompt utilizes the expert pool. Prompts that induce similar internal routing behavior should have similar signatures.

\subsection{Similarity Metric}

We compare routing signatures using mean layer-wise cosine similarity:

\begin{equation}
sim(A,B) =
\frac{1}{L}
\sum_{\ell=1}^{L}
\cos\left(
s_A^{(\ell)},
s_B^{(\ell)}
\right).
\end{equation}

This choice is motivated by three considerations. First, cosine similarity compares activation \emph{patterns} rather than raw counts. Second, layer-wise averaging prevents a single dominant layer from overwhelming the comparison. Third, because signatures are normalized per layer, the metric is robust to differences in prompt length.

\subsection{Expected Similarity Under Random Routing}

If routing decisions were random but balanced across experts, each expert would be selected with roughly equal probability. Under uniform expert sampling, the expected activation distribution becomes approximately

\begin{equation}
s_{\ell,e} \approx \frac{k}{E},
\end{equation}

where $k$ is the number of selected experts per token and $E$ is the total number of experts.

This implies that even random balanced routing can produce moderate similarity between prompts, because activation distributions converge toward uniformity. Therefore, a meaningful routing analysis must compare empirical similarity not only to random permutation baselines but also to a load-balancing baseline.

\subsection{Task-Conditioned Routing}

We formalize the intuition behind our analysis as follows.

\textbf{Proposition 1.}
If hidden-state distributions differ across tasks, then routing distributions differ across tasks.

\textbf{Proof sketch.}
Routing probabilities are computed from hidden states through the router projection and softmax. Hidden-state distributions depend on the statistical structure of prompts. Therefore, if task categories induce systematically different hidden-state distributions, then router outputs will also differ. Consequently, the empirical expert activation frequencies summarized by routing signatures will not be identically distributed across tasks.

This proposition does not require that individual experts correspond to clean semantic skills. It only requires that task-conditioned hidden representations induce distinguishable expert-activation statistics.

\subsection{Information Perspective}

Let $S$ denote routing signatures and $T$ denote task labels. If routing signatures support high-accuracy classification of tasks, then they contain information about task identity, implying

\begin{equation}
I(S;T) > 0.
\end{equation}

We do not estimate mutual information directly in this work, but our classification results provide strong evidence that routing signatures encode statistically extractable task information.

\section{Experimental Setup}

\subsection{Model}

We evaluate OLMoE-1B-7B-0125-Instruct \citep{muennighoff2024olmoe}. This model contains 16 MoE layers, 64 experts per layer, and uses top-$k$ routing with $k=8$. Thus, only 8 of 64 experts are active per token in each MoE layer, corresponding to a sparsity level of 12.5\%.

We selected OLMoE because it is open, practically runnable on commodity hardware, and large enough to exhibit meaningful routing structure while remaining tractable for repeated routing analysis.

\subsection{Prompt Dataset}

We curated 80 prompts across four categories:

\begin{table}[H]
\centering
\caption{Prompt categories used in the experiments.}
\begin{tabular}{lcc}
\toprule
Category & Count & Description \\
\midrule
Code & 20 & Programming tasks and algorithmic prompts \\
Math & 20 & Mathematical and symbolic reasoning prompts \\
Story & 20 & Creative writing and narrative prompts \\
Factual & 20 & Knowledge retrieval and question-answering prompts \\
\bottomrule
\end{tabular}
\end{table}

The categories were selected to span diverse kinds of reasoning and generation behavior. Code and math emphasize structured reasoning. Story emphasizes open-ended generation. Factual prompts emphasize retrieval and explanation.

Each prompt generated 32 tokens during inference.

\subsection{Trace Collection}

Routing traces were collected during inference. For each token we recorded:

\begin{itemize}
    \item layer index
    \item expert index
    \item token position
    \item token type (prompt or generation)
\end{itemize}

All traces were stored as JSONL files. From these traces we computed routing signatures for each prompt.

\subsection{Classifier Setup}

To test whether routing signatures are linearly separable, we trained a logistic regression classifier using routing signatures as input features.

\begin{itemize}
    \item Input dimensionality: 1024
    \item Labels: four task categories
    \item Validation: 5-fold stratified cross-validation
    \item Preprocessing: standardization fit separately on each training fold
\end{itemize}

This setup ensures that classification performance cannot be attributed to data leakage or global normalization artifacts.

\subsection{Baselines}

We evaluate routing similarity relative to two explicit controls:

\paragraph{Permutation baseline.}
Expert assignments are randomly permuted within each layer. This destroys structured routing while preserving sparsity statistics.

\paragraph{Load-balancing baseline.}
Routing is simulated under uniform random expert selection while preserving the empirical per-layer activation totals. This estimates the similarity expected if routing were governed solely by balancing constraints.

\section{Results}

\subsection{Category-Level Routing Similarity}

\begin{figure}[H]
\centering
\includegraphics[width=0.72\linewidth]{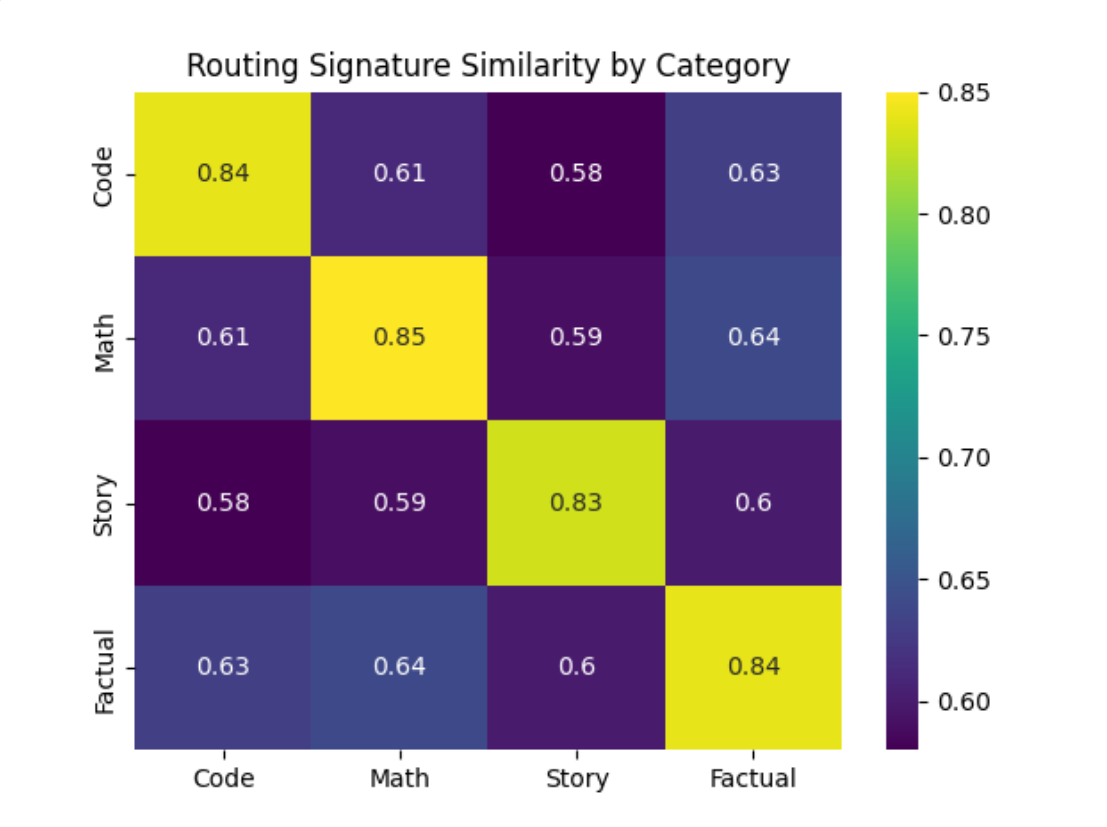}
\caption{Routing signature similarity matrix across task categories. The diagonal entries correspond to within-category similarity and are consistently higher than off-diagonal entries.}
\label{fig:heatmap}
\end{figure}

Figure~\ref{fig:heatmap} shows the routing signature similarity matrix across the four task categories.

The matrix exhibits strong diagonal dominance. Within-category similarities lie between 0.83 and 0.85, while cross-category similarities are substantially lower, typically between 0.58 and 0.64. This pattern indicates that prompts from the same task category activate similar expert distributions, while prompts from different categories activate different subsets of experts.

This is the first indication that routing behavior is task-conditioned rather than random.

\subsection{Baseline Comparison}

\begin{figure}[H]
\centering
\includegraphics[width=0.62\linewidth]{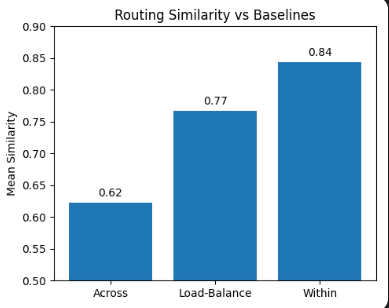}
\caption{Routing similarity compared to the load-balancing baseline. Empirical similarity follows the clean ordering Across $<$ Load-Balance $<$ Within.}
\label{fig:baseline}
\end{figure}

Figure~\ref{fig:baseline} compares mean routing similarity across three conditions: across-category prompt pairs, the load-balancing baseline, and within-category prompt pairs.

We observe the ordering

\begin{equation}
Within > LoadBalance > Across.
\end{equation}

This is important. If routing structure were explained purely by sparse balancing constraints, then empirical similarity would cluster around the load-balancing baseline. Instead, within-task prompts are \emph{more similar} than the baseline predicts, while cross-task prompts are \emph{less similar}. This demonstrates that routing structure exceeds what would arise from balancing alone.

\subsection{Layer-wise Task Signal}

\begin{figure}[H]
\centering
\includegraphics[width=0.62\linewidth]{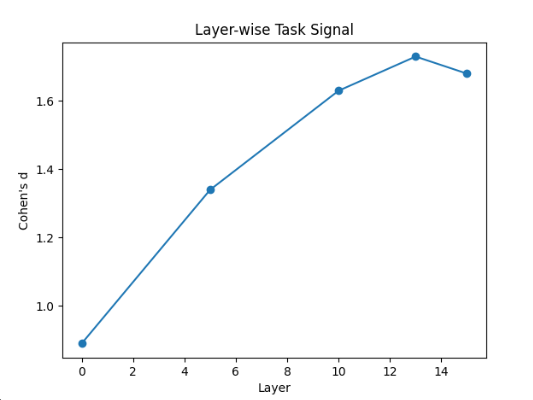}
\caption{Layer-wise task signal strength measured by Cohen's $d$. Task-conditioned separation grows stronger toward deeper layers.}
\label{fig:layer_signal}
\end{figure}

Figure~\ref{fig:layer_signal} shows the layer-wise effect size (Cohen's $d$) separating within-category and across-category routing similarities.

Task separation is weakest in early layers and strongest in deeper layers, peaking around layer 13. This suggests that routing specialization emerges gradually as token representations become more abstract and task-specific. Early layers likely capture lexical and local structure, while deeper layers reflect more semantically differentiated computation.

\subsection{Routing Geometry}

\begin{figure}[H]
\centering
\includegraphics[width=0.72\linewidth]{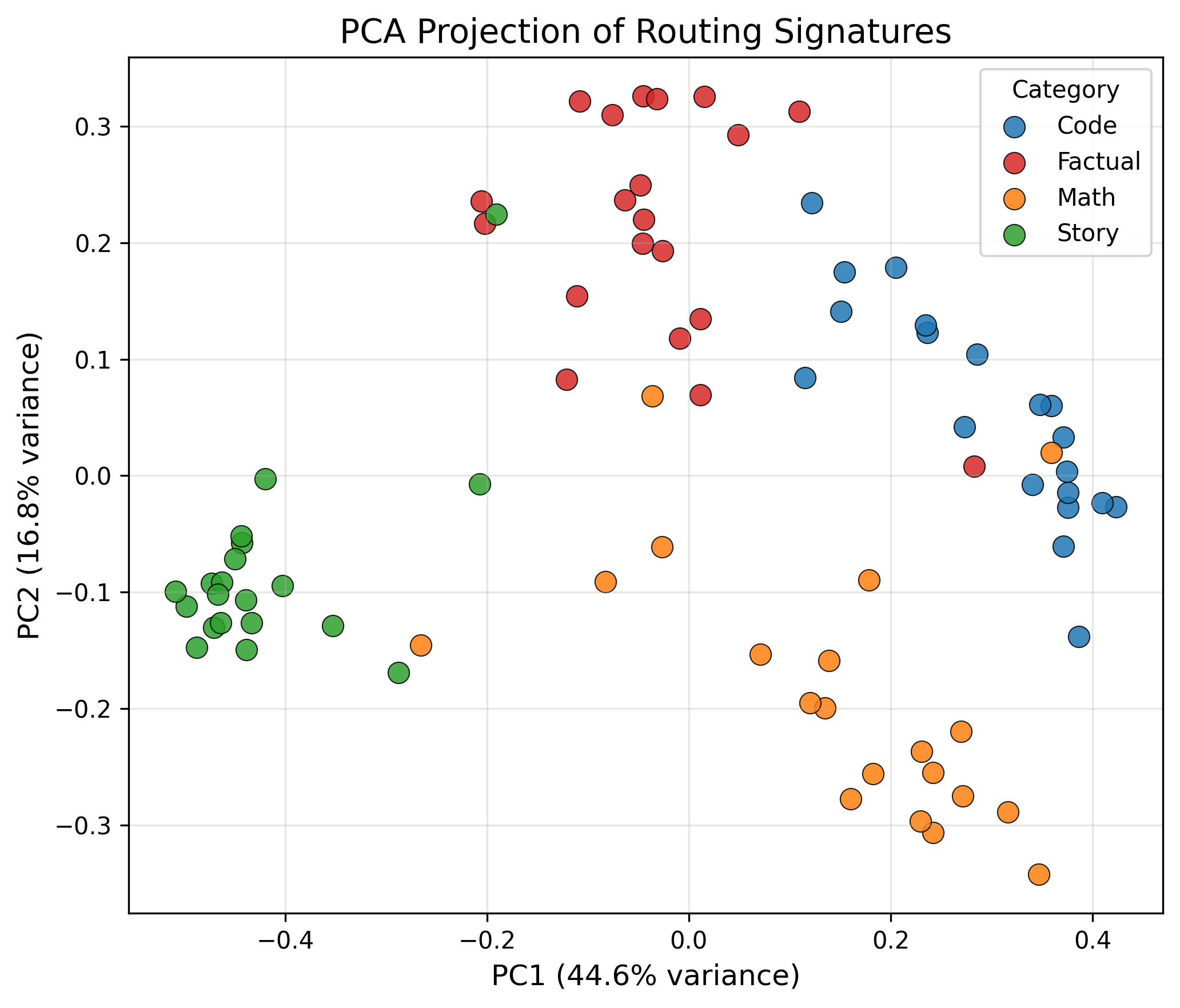}
\caption{PCA projection of routing signatures. Distinct clusters emerge for code, math, story, and factual prompts.}
\label{fig:pca}
\end{figure}

Figure~\ref{fig:pca} visualizes routing signatures projected into two dimensions using PCA.

The first two principal components explain a substantial fraction of the variance, and the projected points form distinct clusters aligned with task categories. Story prompts occupy a clearly separated region, while code and math form different but partially adjacent clusters, consistent with their shared structured reasoning properties. Factual prompts also form a distinguishable cluster.

This visualization does not itself prove separability, but it is consistent with the quantitative classification results reported next.

\subsection{Classification Performance}

A logistic regression classifier trained solely on routing signatures achieves

\begin{equation}
92.5\% \pm 6.1\%
\end{equation}

accuracy across five cross-validation folds, with macro F1 of 0.93.

This result is striking because the classifier sees only routing patterns, not token identities or output text. Therefore, task information is linearly accessible from routing signatures alone.

Taken together with the heatmap, baseline comparison, and PCA projection, this provides strong evidence that routing signatures encode task-discriminative structure.

\section{Discussion}

\subsection{What the Results Show}

Our results support four main claims.

First, routing signatures cluster strongly by task category. Second, this clustering exceeds what would be expected under sparse balancing constraints alone. Third, task signal increases toward deeper layers, suggesting that routing specialization grows with representational depth. Fourth, routing signatures alone are sufficient to support high-accuracy linear classification of task type.

These observations are consistent with the view that MoE routing acts as a conditional computation policy. Different tasks induce different hidden-state distributions, which in turn bias the router toward different subsets of experts.

\subsection{Interpretability Implications}

The main interpretability implication is that routing telemetry provides a direct statistical lens into sparse computation. Unlike weight-space analysis or activation attribution, routing signatures are lightweight and easy to extract. They therefore offer a practical way to study whether sparse models allocate different computation pathways across tasks.

This could be useful for:
\begin{itemize}
    \item expert utilization monitoring,
    \item diagnosing routing collapse,
    \item comparing sparse models,
    \item studying task-conditioned specialization.
\end{itemize}

\subsection{What We Do Not Claim}

It is important to state the limits of our conclusions. We do \emph{not} claim that:
\begin{itemize}
    \item individual experts correspond to clean human-interpretable skills,
    \item routing is the only mechanism responsible for task behavior,
    \item or MoE models implement strict modular cognition.
\end{itemize}

Our claim is narrower and more defensible: routing patterns contain measurable, statistically discriminative task information.

\subsection{Limitations}

This study has several limitations. It evaluates only one model and one prompt dataset. The analysis is correlational rather than causal; we measure routing structure but do not intervene on experts. The dataset is modest in size, and generation length is limited. These constraints make the paper a focused empirical note rather than a comprehensive theory of MoE routing.

\subsection{Future Work}

Several natural next steps follow from this work:

\begin{itemize}
    \item cross-model comparison of routing signatures,
    \item longer-horizon routing dynamics across extended generation,
    \item causal intervention on experts,
    \item routing-aware decoding or task adaptation,
    \item direct estimation of mutual information between routing and task.
\end{itemize}

\section{Conclusion}

We introduced routing signatures as a framework for analyzing expert routing in sparse MoE transformers. Using OLMoE as a testbed, we showed that routing signatures cluster by task category, exceed load-balancing expectations, strengthen in deeper layers, and support accurate linear task classification. These findings suggest that routing is not merely a balancing mechanism but a measurable task-sensitive component of sparse transformer computation.

\bibliographystyle{unsrtnat}
\bibliography{references}

\appendix

\section{t-SNE Visualization}

\begin{figure}[H]
\centering
\includegraphics[width=0.72\linewidth]{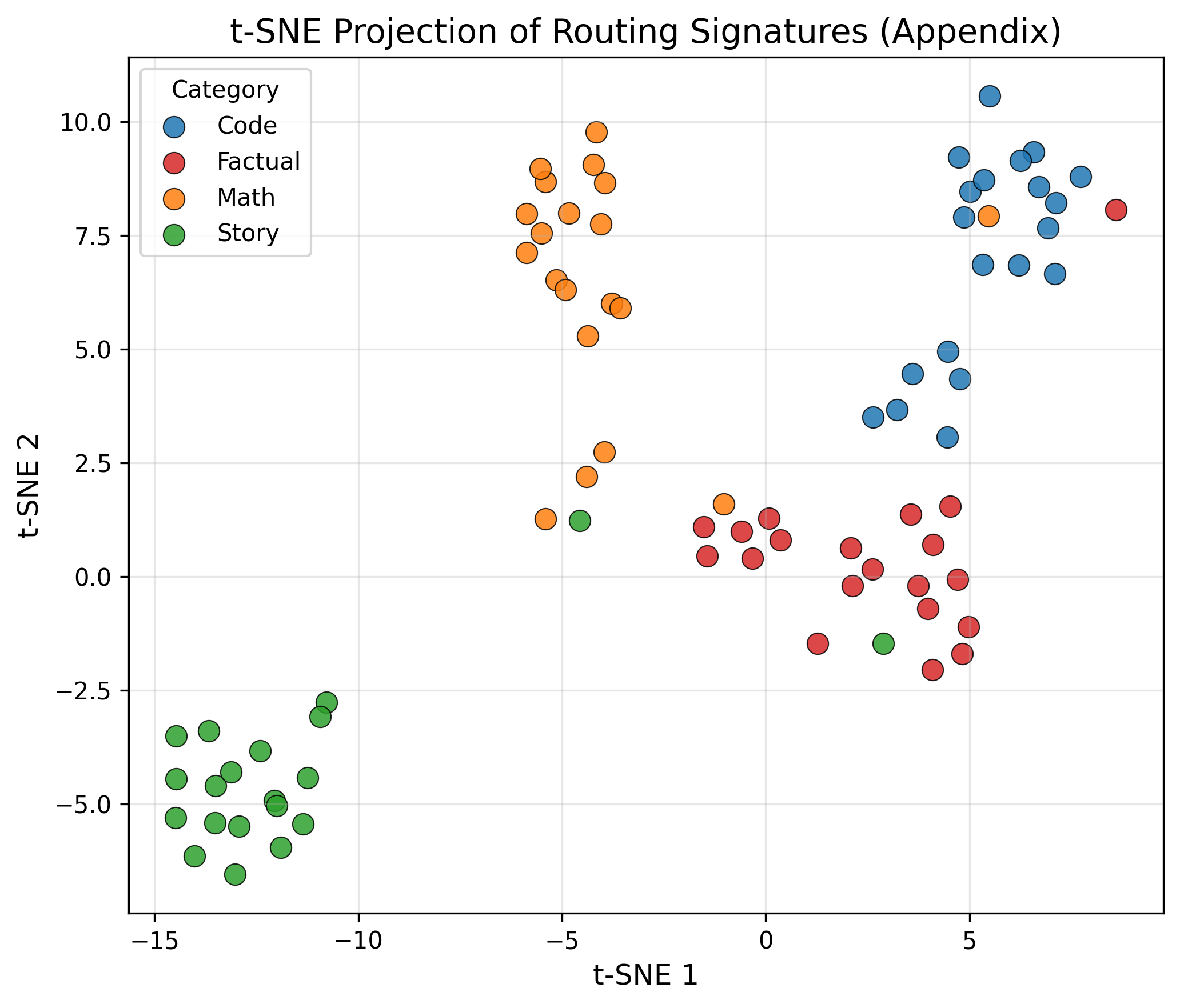}
\caption{t-SNE projection of routing signatures showing clear task-conditioned clusters.}
\label{fig:tsne}
\end{figure}

Figure~\ref{fig:tsne} shows a t-SNE projection of routing signatures. Unlike PCA, t-SNE emphasizes local neighborhood structure. The resulting plot shows clear category-level grouping, qualitatively reinforcing the main analysis.

\end{document}